\documentclass{article}

\usepackage{arxiv}

\usepackage[utf8]{inputenc} 
\usepackage[T1]{fontenc}    
\usepackage{hyperref}       
\usepackage{url}            
\usepackage{booktabs}       
\usepackage{amsfonts}       
\usepackage{nicefrac}       
\usepackage{microtype}      
\usepackage{lipsum}		
\usepackage{graphicx}
\usepackage[numbers]{natbib}
\usepackage{doi}
\usepackage{xcolor}

\title{A universal synthetic dataset for machine learning on spectroscopic data}

\author{ 
	\href{https://orcid.org/0000-0001-5816-4866}{\includegraphics[scale=0.06]{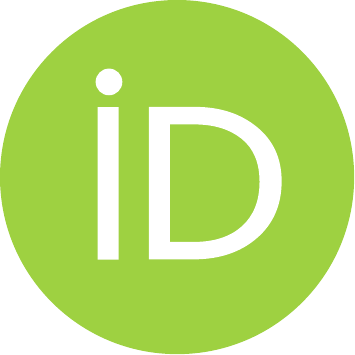}\hspace{1mm}Jan Schuetzke} \\
	Institute for Automation and Applied Informatics\\
	Karlsruhe Institute of Technology\\
	Karlsruhe, Germany \\
	\texttt{jan.schuetzke@kit.edu} \\
	\And
	\href{https://orcid.org/0000-0003-2255-9676}{\includegraphics[scale=0.06]{orcid.pdf}\hspace{1mm}Nathan J.~Szymanski} \\
	Department of Materials Science \& Engineering \\
	UC Berkeley, Lawrence Berkeley National Laboratory\\
	Berkeley, CA 94720, USA \\
	\texttt{nathan\_szymanski@berkeley.edu} \\
	\And 
	\href{https://orcid.org/0000-0002-7780-6374}{\includegraphics[scale=0.06]{orcid.pdf}\hspace{1mm}Markus Reischl} \\
	Institute for Automation and Applied Informatics\\
	Karlsruhe Institute of Technology\\
	Karlsruhe, Germany \\
	\texttt{markus.reischl@kit.edu} \\
}

\renewcommand{\shorttitle}{Synthetic Spectroscopic Dataset for Machine Learning}

\hypersetup{
pdftitle={A universal synthetic dataset for machine learning on spectroscopic data},
pdfsubject={Deep Learning, Spectroscopic Data, XRD, NMR, Raman},
pdfauthor={Jan Schuetzke},
pdfkeywords={Deep Learning, Spectroscopic Data, Benchmarking},
}

\begin{document}
\maketitle

\begin{abstract}
    To assist in the development of machine learning methods for automated classification of spectroscopic data, we have generated a universal synthetic dataset that can be used for model validation. This dataset contains artificial spectra designed to represent experimental measurements from techniques including X-ray diffraction, nuclear magnetic resonance, and Raman spectroscopy. The dataset generation process features customizable parameters, such as scan length and peak count, which can be adjusted to fit the problem at hand. As an initial benchmark, we simulated a dataset containing 35,000 spectra based on 500 unique classes. To automate the classification of this data, eight different machine learning architectures were evaluated. From the results, we shed light on which factors are most critical to achieve optimal performance for the classification task. The scripts used to generate synthetic spectra, as well as our benchmark dataset and evaluation routines, are made publicly available to aid in the development of improved machine learning models for spectroscopic analysis.
\end{abstract}


\section{Introduction}
\label{sec:introduction}

Spectroscopic techniques such as X-ray diffraction (XRD), Nuclear Magnetic Resonance (NMR) and Raman scattering are fundamental tools for the characterization of experimental samples in chemistry and materials science. XRD has been employed throughout industry and research laboratories for more than a century \cite{laue1913}, and is well suited to characterize crystalline materials as it captures detailed information regarding the long-range, periodic nature of their structures. NMR and Raman measurements, on the other hand, depend more strongly on localized chemical interactions and are widely used to characterize the structure of molecular materials \cite{ernst1987, smith2019}. While their mechanisms and applications may differ, each of these characterization techniques produce similar one-dimensional spectra (sometimes referred to as \textit{patterns}) containing peaks with distinct positions, widths, and intensities. These features often serve as “fingerprints” for molecules and crystalline phases, which can be used to match unknown samples. Phase identification can be accomplished by comparing newly measured spectra with those of previously reported materials in experimental databases such as the ICSD or RRUFF \cite{icsd, rruff}. To automate this process, machine learning has recently emerged as an effective tool that can map experimental spectra onto known structures, with reported accuracies exceeding standard similarity-based metrics \cite{choudhary2022, szymanski2021a}.

Machine learning models based on neural networks have been most widely used for spectroscopic analysis. This was first demonstrated in the work of Park et al., where a convolutional neural network was trained to classify XRD patterns by their structural symmetries (i.e., space groups) \cite{park2017}. Later work extended the use of neural networks to identify particular phases from XRD patterns, even dealing with multi-phase mixtures \cite{lee2020, schuetzke2021, szymanski2021b}. For the analysis of NMR and Raman spectra, similar methods have also been used to aid manual analysis \cite{nmrreview2020} and automate the identification of molecular species \cite{raman1, raman2, raman3}. In each of these studies, a unique network architecture was developed and optimized to perform well for a specific characterization technique, though it remains in question whether a single generalized architecture could be applied to handle all measurement types.

Neural networks are known to require large amounts of data for training and validation. There exists a variety of databases containing experimental spectroscopic data for materials and molecules. For example, the RRUFF provides a mix of XRD and Raman spectra for about 5,800 minerals \cite{rruff}. Similarly, NMRShiftDB provides NMR spectra for about 45,000 organic molecules \cite{nmrshiftdb2}. In general, however, these databases cover only a small portion of the entire chemical space (e.g., >260,000 known crystal structures exist in the ICSD) \cite{icsd}. Moreover, because spectral data can vary between different measurements due to sample artifacts and instrumental aberrations, the reported spectra may not accurately represent all future results. For example, XRD patterns often show differences in peak positions and heights due to strain and preferred orientation in the sample. Likewise, changing the buffer solution used during NMR measurements can lead to noticeably different spectra for identical molecules.

To overcome the limited availability of diverse experimental data, past work has instead used large sets of simulated spectra for the training and validation of machine learning models designed to analyze such data \cite{schuetzke2021, szymanski2021b}. On the one hand, XRD patterns can be rapidly calculated through a Fourier transform on a given crystal structure \cite{pecharesky2005}. On the other hand, accurate simulation of Raman and NMR spectra requires more costly ab initio calculations based on density functional theory \cite{persson_raman, nmr_sim}, limiting the rate at which training data can be generated for machine learning.

Here, we develop a completely universal synthetic dataset that can be used to train and validate models for the analysis of spectroscopic data. The dataset contains features that are shared between distinct characterization techniques, thereby providing a representation of spectra obtained by XRD, Raman, or NMR measurements. Because this algorithm does not rely on physics-based simulations, it rapidly generates data with little computational cost. A total of 100,000 spectra were simulated in about 30 seconds using a standard desktop computer. Furthermore, because this dataset is not explicitly derived from known materials or molecules, it excludes any experimental bias and is readily tailored to suit specific test cases (e.g., to probe the effects of peak overlap). We hope that this work will provide a benchmark for classification models on spectral data. To this end, we apply several previously reported machine learning architectures to the simulated dataset and report their accuracies. All code and data discussed here is made publicly available, and we invite the community to contribute to the existing results.

\section{Methods}
\label{sec:methods}

\begin{figure}[t]
	\centering
	\includegraphics[width=\textwidth]{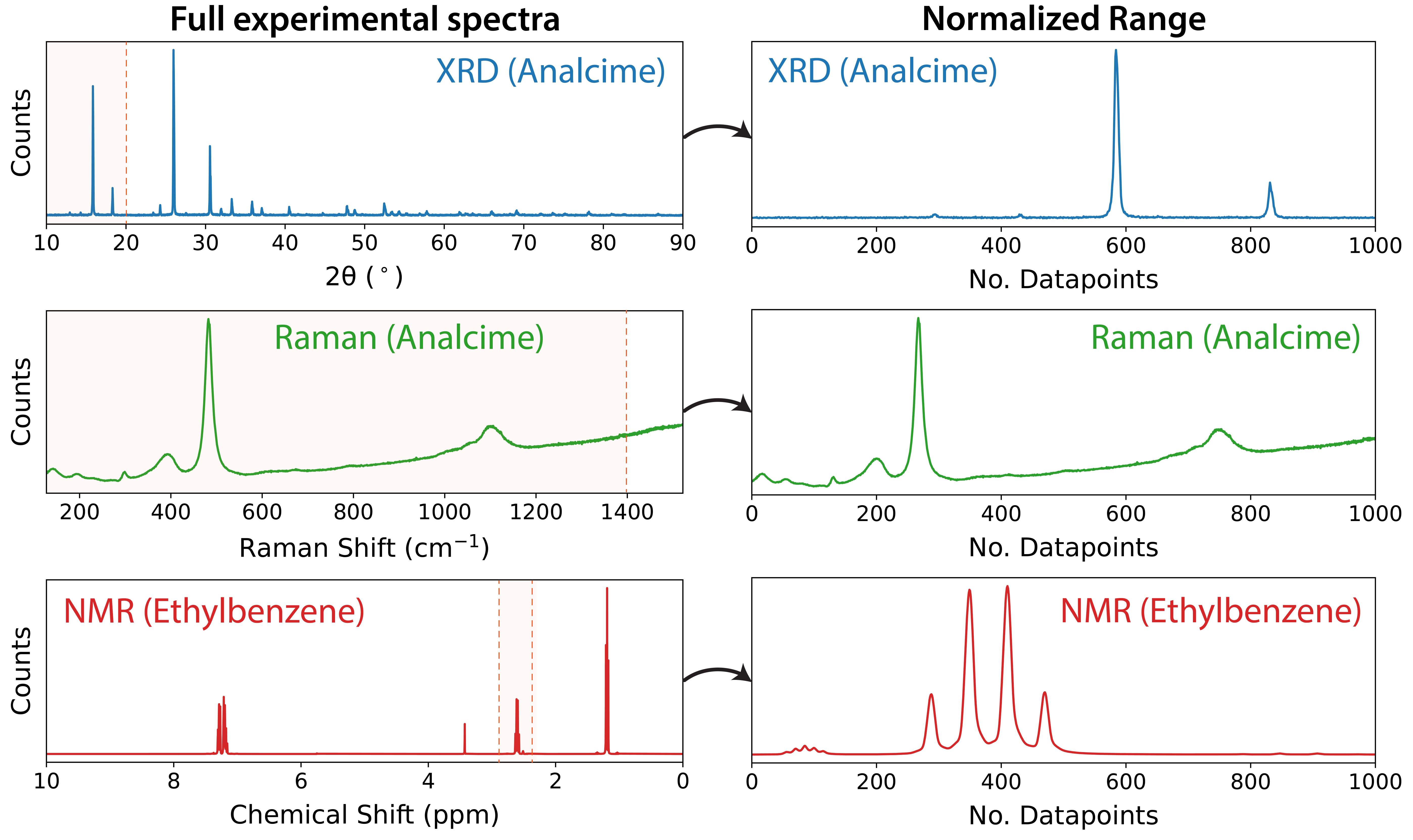}
	\caption{Three examples of spectral data are shown. XRD and Raman patterns are given for Analcime, a common inorganic mineral \cite{rruff}. An NMR spectrum is presented for Ethlybenzene, a well-known organic compound \cite{nmrium}. The left panels shows raw data spanning the entire measurement range. The right panel shows the same spectral data, now standardized to contain 1000 measurement datapoints. The corresponding ranges are highlighted (orange) in the left panel.}
	\label{fig:exemplary-spectra}
\end{figure}

Figure \ref{fig:exemplary-spectra} shows three exemplary spectra obtained from XRD, Raman and NMR measurements on samples of Analcime and Ethylbenzene.
As shown in the left panel of Figure \ref{fig:exemplary-spectra}, spectra obtained using different techniques appear quite different at first glance. For example, Raman spectra often contain fewer and more diffuse peaks than XRD of NMR patterns. However, upon further inspection, the density of peaks with respect to the sampling of datapoints remains relatively consistent between different techniques. The Raman pattern shown in Figure \ref{fig:exemplary-spectra} contains only 1000 distinct datapoints, whereas the XRD and NMR patterns contain 8000 and 65000 datapoints, respectively. If we instead crop each of the spectra to contain only 1000 datapoints, as shown in the right panel of Figure \ref{fig:exemplary-spectra}, the resulting patterns appear much more similar. Regardless of which technique is used, the normalized spectra each contain 2-6 peaks with comparable widths. Therefore, we conclude that each technique produces data with similar "information density," justifying our choice to use a single artificial dataset for testing and validation of machine learning models.

For all characterization techniques considered here, experimental artifacts and instrumental aberrations can cause spectra to deviate from their expected fingerprints.
For example, peak intensities from XRD scans often show large changes caused by a non-random orientation of particles in the specimen.
Similarly, the positions of peaks in NMR spectra can be shifted by the choice of buffer solution used during measurement. 
To account for such effects, our synthetic dataset introduces stochastic variations related to changes in peak positions, intensities and widths.
However, these variations should be kept minimal as to avoid unwanted overlap between specimen with similar fingerprints.
Experimentally, the reduction of artifacts can be accomplished through careful sample preparation (e.g., grinding of powder samples for XRD). 
Here, we limit the variations included in our synthetic dataset such that all spectra could, in theory, be successfully classified by a human expert.

\newpage
In the following Section \ref{sec:synthetic-dataset}, we outline a general approach to simulate spectra with customizable features. 
Using this approach, we generate a large synthetic dataset based on the parameters detailed in Section \ref{sec:benchmark-dataset}. Several neural network architectures are tested on this set, as described in Section \ref{sec:network-structures} and \ref{sec:reproducability}
All code used here can be found in the github repository \href{https://github.com/jschuetzke/synthetic-spectra-generation}{jschuetzke/synthetic-spectra-generation}.

\subsection{Synthetic Dataset Approach}
\label{sec:synthetic-dataset}

In order to generate a consistent spectroscopic dataset, several parameters must be specified.
For example, every spectrum should contain an equal number of datapoints that span a fixed range (i.e., equal start and end points). 
Additionally, there should be a fixed number of possible classes used to generate unique spectra, as these classes will be used to train machine learning models later on. 
Accordingly, we define a set of universal parameters for the synthetic dataset as follows:

\begin{itemize}
    \item The number of datapoints contained in each pattern
    \item The number unique classes in the dataset (ideal patterns)
    \item The minimum and maximum number of peaks in each pattern
\end{itemize}

In contrast to experimental scans, our synthetic spectra do not have a real measurement range but are instead described by the number of datapoints contained by each spectrum.
To simulate spectra that are representative of experimental data (i.e., similar to the spectra shown in Figure \ref{fig:exemplary-spectra}), we recommend that parameters are chosen to produce a comparable "information density", e.g., with 2 to 6 peaks per 1000 datapoints.
The parameters can be manually set by the user to fit the problem at hand in (\href{https://github.com/jschuetzke/synthetic-spectra-generation/blob/main/dataset_config_generator.py}{\textit{dataset\_config\_generator}}).
Based on the settings, we simulate classes by randomly sampling different peak positions and relative intensities.
Following the execution of the generator script, each class is defined by a unique set of discrete peak positions and intensities. This class information is stored in single file describing the entire dataset.

We next generate spectra from the ideal class representations by adding minor alterations that account for possible experimental artifacts.
The position and intensity of each peak is randomly varied, and the resulting information is used to fit a Guassian bell curve. These curves are then summed to form a continuous spectrum. 
This process relies on the following parameters:

\begin{itemize}
    \item The number of augmented patterns generated for each class
    \item The magnitude of applied variations:
    \begin{itemize}
        \item Maximum peak position shift,
        \item Maximum peak intensity change,
        \item Range of Gaussian peak widths.
    \end{itemize}
\end{itemize}

Here, all variations are applied independently of one another. This differs from experimental artifacts, where changes are often coupled. For example, strain-induced changes in the peak positions of XRD spectra are known to follow a well-defined relation. Moreover, peak shapes are often more complex than simple Gaussian bell curves (e.g., Voigt profiles), and the widths of these peaks can be correlated with one another throughout a given spectrum.
However, our current approach is designed with simplicity and generalizability in mind. The resulting data should still be sufficient for the validation of machine learning models, which can be applied to experimental data later on.
A key advantage of simulations is their ability to rapidly generate customized synthetic datasets, which can be used to test classification algorithms with respect to different features and variations.
For example, one can evaluate how many randomly varied samples are necessary per class to train a robust classification model given certain parameters (number of datapoints, peak counts, and degree of overlap).
The code used to simulate continuous spectra can be found in the script, \href{https://github.com/jschuetzke/synthetic-spectra-generation/blob/main/spectra_from_config.py}{\textit{spectra\_from\_config}}.


\subsection{Robust Classification Benchmark Dataset}
\label{sec:benchmark-dataset}

\begin{figure}[t]
	\centering
	\includegraphics[width=\textwidth]{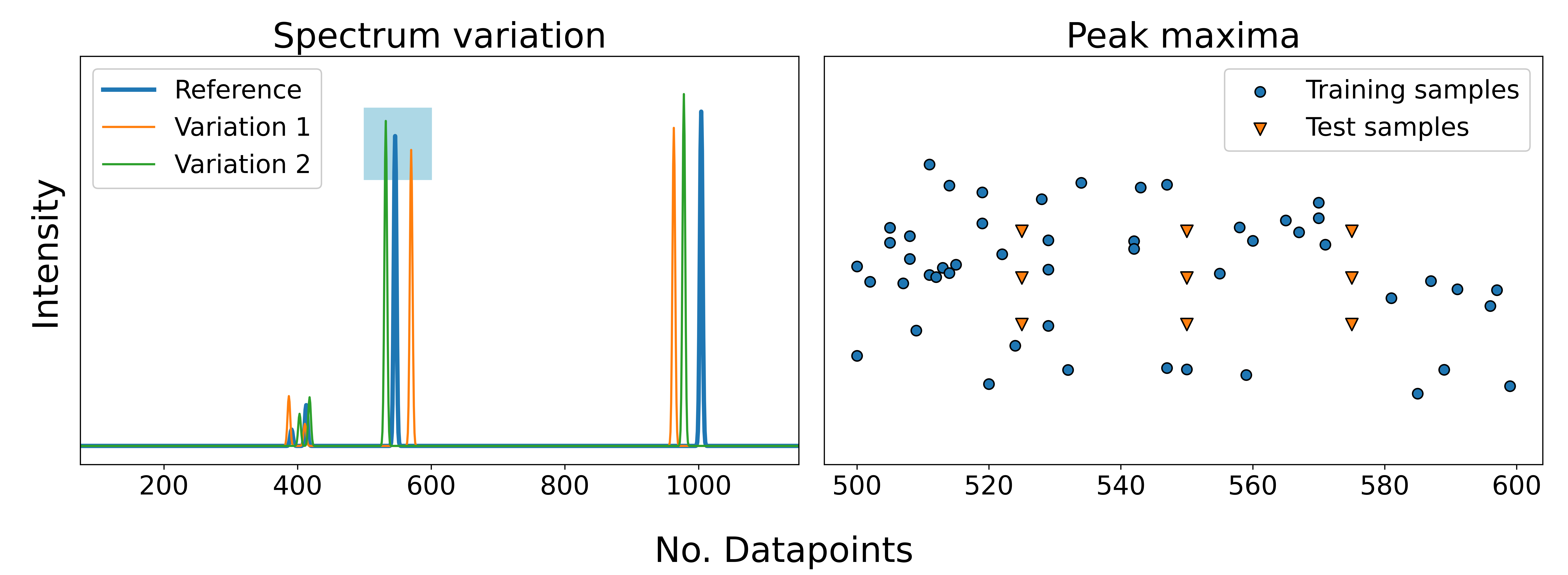}
	\caption{An illustration of the variations included in our spectra dataset. The left pane shows three varied spectra of a class, in with the defined set of peaks with discrete position and intensity information is transformed into continuous spectra. The blue line shows the reference pattern in which the peak information is transformed without any variations. For the orange and green graph, the positions and intensities are slightly shifted to account for realistic deviations. Second, we show the maximum value of the second peak in detail (as outlined by the opaque blue rectangle) in the right pane. Here, we show examples for the position and heights of randomly varied training samples (blue), as well as structurally varied test samples for evaluation purposes.}
	\label{fig:class_variation}
\end{figure}

To demonstrate the benefits of our simulation algorithm, we show how a dataset can be generated and used to benchmark several different neural network models designed for the classification of spectroscopic data.
These tests will be used to determine whether more complex models achieve better classification accuracy than "simpler" algorithms with shorter training times.
To this end, we design a dataset that has some minor overlap in the training data, but little to no overlap in the test data such that all samples should be correctly classified by a perfect model.
With this dataset, we investigate how well each model is able to learn generalized representations for the classes while ignoring the inconclusive training samples from overlapping regions.

By testing various configurations of our dataset generator, we identified a set of parameters that include a large number of classes (500) and only slight overlap between similar spectra when variations are applied.
With these parameters, each pattern is comprised of 5000 datapoints that contain 2 - 10 peaks.
We also impose limits on all peak positions such that they maintain a distance of at least 100 datapoints from the borders of the spectrum. This ensures that no peaks are lost when their positions are varied.
Based on the parameters used here, a json file is generated containing the information associated with each class. This file can be found in our repository: 
 \href{https://github.com/jschuetzke/synthetic-spectra-generation/blob/main/dataset_configs/dataset500.json}{\textit{dataset\_configs/dataset500.json}}. 

To accurately measure the performance of the machine learning models tested here, we generate separate training and test sets, each of which are structurally different from one another to avoid data leakage.
A perfect model should not only perform well on the data used during training, but also on previously unseen samples.
To form the test set, we include ideal spectra without any variations to their peak positions and intensities, as well as narrow peak widths (kernel size 2). Additionally, we include spectra with a uniform grid of variations. Peak positions are shifted by 25 datapoints in either direction, and peak heights are varied by +/- 2\%. By combining these variations, we form a test set that contains 9 samples per class (ideal spectrum + 8 alterations). 
For comparison, 60 randomly generated samples are used for training and validation, containing larger variations with shifts as large as 50 datapoints, +/- 5\% changes in peak intensities, and kernel sizes from 2 - 5. 
The variations included in the test set are intentionally weaker than those included during training to ensure that no overlap occurs in the test set. 

The effects of these variations are illustrated in Figure \ref{fig:class_variation}.
The left pane shows three continuous spectra, including the ideal representation (blue) of the specified class, as well as two variations (green and orange).
The variations illustrate independent changes in the peak positions and intensities, relative to the ideal spectrum.
To better visualize such variations, the right pane shows discrete peak information (position and height) from samples in the training and test set. These represent peaks from the blue-shaded region in the left pane.
As shown, the training samples (blue circles) are distributed randomly across a broad range of positions and heights, whereas the test samples (orange triangles) follow a uniform arrangement with less variation.

All patterns included in our benchmark dataset have been saved as numpy arrays (x.npy), complemented with the corresponding class affiliation (y.npy). These files are both available in:  \href{https://osf.io/pqahd/?view_only=5d31d8c981aa4b5faa299fa79ba49f0d}{\textit{osf.io}}.

\subsection{Neural Network Architectures}
\label{sec:network-structures}
Spectroscopic data can be classified in a number of ways. 
For example, several models have been developed to categorize the space group of crystalline materials from XRD patterns, thereby grouping completely different patterns into a single class defined by symmetry relations among peaks \cite{park2017,dong2021,oviedo2018}. 
In contrast, it is more common to use spectral data for the identification of distinct classes by comparing the positions and heights of observed peaks with known reference data. 
This is the task that we consider in this work, which may be framed as a one-to-one mapping between each "fingerprint" and class. 
Accordingly, we only test machine learning architectures that have been developed for this purpose, including models designed for the identification of distinct phases or chemical spaceies from XRD patterns and Raman spectra \cite{raman1,raman2,raman3}.

All models tested employ a neural network architecture containing the following components:

\begin{enumerate}
    \item Convolutional layers trained to extract local features that are independent of position and orientation.
    \item Fully-connected layers that learn classification rules based on the previously extracted features.
\end{enumerate}

While the models all follow this general architecture, their exact implementation varies from paper to paper. Here, we test eight different models with architectures summarized in \ref{tab:network-structures}.
The first three networks employ a straightforward structure containing stacked convolutional layers without any extra modifications (e.g. batch normalization), followed by two fully connected layers.
All three plain CNNs apply Maximum-pooling (maxpool) operations following the convolutional layers to reduce the dimensionality of the input scan.
We further test more sophisticated networks that use batch-normalization between convolutional layers and activation functions (CNN BN), as well as a VGG-like architecture that stacks even more convolutional layers before the pooling operation (VGG).
An important concept commonly found in neural networks applied to image data are residual blocks (Resnet) that use a large amount of convolutional layers to extract complex features, which is only possible by using "residual connections" that prevent the vanishing gradient problem in deep neural networks.
Similarly, inception blocks (INC3 \& INC6) combine different kernel sizes in convolutional blocks to account for different feature sizes.

We benchmark the different model implementations as they are presented in their respective papers, without any modifications.
Each model is trained using an Adam classifier with default hyperparameters ($\beta_1=0.9$, $\beta_2=0.999$) and a learning rate of $3e^{-4}$, with a mini-batch size of 128.
Since different model architectures require more or less time to converge, we set a high number of training epochs (500) and use the Early-Stopping technique to halt training once the validation loss levels off.
Additionally, we lower the learning rate by a factor of 0.5 if the validation loss does not improve for 10 consecutive epochs, while early stopping activates at 25 epochs without any improvement.

\begin{table}[t]
\caption{A description of the neural network architectures tested in this work. The network structure is described with \textit{C} for 1D convolutional layers, \textit{MP} for 1D Max Pooling operations and \textit{BN} for batch normalization. More complex structures like residual (RES) and inception (INC) blocks are only abbreviated here. The full details, like the kernel sizes for the convolutions, can be found in their respective publications or in the repository \href{https://github.com/jschuetzke/synthetic-spectra-generation/tree/main/model_implementations}{jschuetzke/synthetic-spectra-generation/model\_implementations}.}
\centering
\begin{tabular}{lllll}
Model Name & Publication & Architecture & No. Neurons & Conv-Output \\ \hline
CNN2 & \citet{lee2020} & (C-MP)x2 & 2000-500 & 139x64 \\
CNN3 & \citet{lee2020} & (C-MP)x3 & 2500-1000 & 139x64 \\
CNN6 & \citet{szymanski2021b} & (C-MP)x6 & 3100-1200 & 78x64 \\
VGG  & \citet{wang2020} & C-MP-(C-C-MP)x3 & 120-84-186 & 312x64 \\
CNN BN & \citet{raman1} & (C-BN-MP)x3 & 2048 & 625x64 \\
Resnet & \citet{raman2} & C-BN-RESx6 & - & 79x100 \\
INC3 & \citet{lee2020} & (C-MP)x3-INCx3 & 2500-1000 & 139x147 \\
INC6 & \citet{lee2020} & (C-MP)x3-INCx6 & 4000-400 & 139x332 \\
\end{tabular}
\label{tab:network-structures}
\end{table}

\subsection{Reproducability}
\label{sec:reproducability}

During our tests, we found that the network performance varied substantially depending on which random seed was used to initialize the training of each model.
Therefore, we aim to benchmark not only the accuracy of each model, but also how much variance is present between different training iterations.
The variation is further complicated by changes that can be introduced when training on GPU machines that utilize CUDA algorithms, which are not deterministic.
Accordingly, the following policies are used to ensure reproducibility.

\textbf{\underline{Training in Docker containers}} Since the training of neural networks in a Python environment requires multiple libraries that are available in different versions, and we cannot confirm that every single combination produces identical results, we train our models in a Docker container.
A Docker container is a small virtual machine that is built according to a recipe (image) and asserts platform-independent performance (Host machine can be Windows, MacOS or Linux).
The container still has a few connections to the host machine (e.g. GPU drivers) that may be subject to change and could cause minor performance differences, but to our knowledge this is the most consistent option.

\textbf{\underline{Setting deterministic algorithm flags}} As mentioned before, TensorFlow (and other libraries like pyTorch) relies on CUDA algorithms that are not fully deterministic by default.
Here, we set tensorflow to only use deterministic operations which hampers the computation speed but assures consistent results.

\textbf{\underline{Multiple model initializations}} We train each model 5 times to investigate performance differences between training runs. 
To account for random variations, we randomly set the seed at the start of each run, which how the model is initialized. We also shuffle the order of the training data between epochs.
Using identical parameters such as the mini-batch size and random seed across models means that each network sees the exact same training samples per step and convergence only depends on initialization and the network's ability to learn meaningful features for the data. 
\section{Benchmark Results}
\label{sec:results}

As presented in Table \ref{tab:basis-benchmark}, all models achieve a high accuracy on our synthetic dataset. 
Even the worst performing model correctly predicts 4370 of the 4500 total test samples (97\% accuracy).
To compare these models in a clear fashion, we report the number of misclassifications made by each model.
Overall, the CNN6 model by \citet{szymanski2021b} achieves the best performance and only misclassifies 12 out of 4500 samples for the best initialization.
Most of the models misclassify about 30 of the test samples, with the exception of the CNN BN, which misclassifies 95 samples.

\begin{table}[t]
\caption{Performance of the published models on the synthetic dataset with 500 classes. Each model is trained 5 times with differing starting initializations (random seeds). We report the absolute numbers of misclassified test samples, the required number of training epochs and time for each model. We achieve this training time using a Titan RTX GPU.}
\centering
\begin{tabular}{lrrrr}
Model Name & Misclassifications & Trained Epochs & Time (min) \\ \hline
CNN2 & 30 +/- 2 & 60-90 & 8\\
CNN3 & 33 +/- 10 & 50-80 & 10\\
CNN6 & \textbf{14 +/- 2} & 75-90 & 30\\
CNN BN & 95 +/- 35 & 35-50 & 6\\
VGG  & 30 +/- 10 & 55-70 & 5\\
Resnet & 25 +/- 7 & 30-35 & 35\\
INC3 & 32 +/- 9 & 55-75 & 16\\
INC6 & 27 +/- 7 & 60-65 & 24\\

\end{tabular}
\label{tab:basis-benchmark}
\end{table}

These tests reveal several interesting trends. 
First, we find that the best performing models are the ones that significantly reduce the dimensionality of the input layer. 
For example, the top two models (CNN6 and ResNet) contain only 78 or 79 values per channel after convolution. 
In contrast, the worst performing models retain a high dimensionality after convolution (e.g., 64 channels each containing 625 values for CNN BN). 
These results suggest that a low dimensionality is preferred to achieve good performance for the classification of spectral data by a neural network. 

Second, our results demonstrate that more complex architectures do not necessarily lead to better performance. 
For example, the resnet and inception networks do not produce less misclassifications than the respective networks without those blocks (INC3 vs. CNN3). 
An inception block is typically used to extract features on a different scale but the inception networks do not show any meaningful performance benefit in comparison to the plain CNNs while taking more time to train.
Likewise, the addition of batch-normalization or the VGG-like architecture yields no improvement.
One significant effect of the more complex architecture is the increased time to train the models, as "simple" models converge in about 5 minutes, while more complex architectures take up to 35 minute training time.
The training times we report apply for a machine with a Titan RTX GPU and are significantly higher when the training is performed on a CPU machine.

Third, we find that there is a large variation in accuracy when using the same architecture trained using different seeds for initialization. 
While each model produces around 30 misclassifications on average, this can vary by +/- 10 in either direction depending on which seed is used. This means that random variations due to initialization or order of training samples can account for as much as 33\% of the accuracy difference. 
The performance variation applies across all models, including the best (CNN6) and worst model (CNN BN). 
These results suggest that the convergence of the training run is heavily dependent on the initialized weights of each model, and the current approaches used for initialization do not result in robust model training.


\section{Conclusion}
\label{sec:conclusion}

To benchmark different neural network architectures for the classification of spectroscopic data, we have provided a method to rapidly simulate artificial spectra with user-defined features. 
Using this method, we generated a dataset that depicts common artifacts related to changes in peak positions, widths, and heights. These changes are designed to represent experimental spectra from different measurement techniques including XRD, NMR, and Raman spectroscopy.
On this dataset, several recently reported models were tested. The results demonstrate that the best-performing models are those that significantly reduce the dimensionality of the input data, thereby simplifying the features that need to be learned by the neural network. 
Furthermore, we find that more advanced architectures (with resnet, inception blocks, or batch normalization) yield no meaningful benefits for the classification of spectroscopic data.  
While all models tested here perform relatively well, there is a large amount of stochastic variations present in the training process, which can lead to noticeably different accuracy depending on which seed is used for initialization. We therefore recommend that authors publish not only the accuracy of their best model, but also the standard deviation of this metric across multiple training runs with different seeds.

We acknowledge that the network architecture presented in their respective publications is usually reported after a hyperparameter-search for their specific data, meaning that the adaption of few parameters could result in a better performance on our dataset.
Nonetheless, we designed our synthetic dataset to accurately model the general characteristics of spectroscopic patterns, so our dataset provides an opportunity to quickly benchmark different network architecture, since we publish all of our scripts and datasets in the specified repositories.
As the ImageNet challenge sparked innovation for the analysis of image data with new architectures such as Resnet, we invite researchers to develop improved models that can outperform existing methods on the dataset presented in this work.

\section*{Acknowledgements}
We want to thank the Karlsruhe House of Young Scientists (KHYS) for funding the scientific exchange that was essential for obtaining the results.

\bibliographystyle{unsrtnat}

\bibliography{references}

\end{document}